\def\year{2022}\relax
\documentclass[letterpaper]{article} 
\usepackage{aaai22}  
\usepackage{times}  
\usepackage{helvet}  
\usepackage{courier}  
\usepackage[hyphens]{url}  
\usepackage{graphicx} 
\usepackage{natbib}  
\usepackage{caption} 
\DeclareCaptionStyle{ruled}{labelfont=normalfont,labelsep=colon,strut=off} 
\usepackage{algorithm}
\usepackage{algorithmic}
\usepackage{amsfonts}
\usepackage{xspace}
\usepackage{enumitem}
\usepackage{array}
\usepackage{multirow}
\usepackage{subcaption}
\usepackage{tablefootnote}
\usepackage{balance}
\usepackage[export]{adjustbox}
\usepackage{xcolor}
\usepackage{esvect}
\usepackage{balance}
\usepackage{times}

\usepackage{soul}
\usepackage[hidelinks]{hyperref}
\usepackage[utf8]{inputenc}
\usepackage{amsmath}
\usepackage{booktabs}
\graphicspath{ {Figures/} }

\title{tBDFS: Temporal Graph Neural Network Leveraging DFS}
\author {
    Uriel Singer$^1$,
    Haggai Roitman$^2$,
    Ido Guy$^3$,
    Kira Radinsky$^1$
}
\affiliations {
    $^1$Technion, Israel Institute of Technology,
    $^2$eBay Research,
    $^3$Ben-Gurion University of the Negev\\
    urielsinger@cs.technion.ac.il,
    hroitman@ebay.com,
    idoguy@acm.com,
    kirar@cs.technion.ac.il
}

\begin{document}

\maketitle

\newcommand{\specialcell}[2][c]{%
  \begin{tabular}[#1]{@{}c@{}}#2\end{tabular}}

\newcommand{\us}[1]{\textcolor{red}{$\ll$\textsf{#1 --US}$\gg$}}
\newcommand{\hr}[1]{\textcolor{blue}{$\ll$\textsf{#1 --HR}$\gg$}}
\newcommand{\ig}[1]{\textcolor{green}{$\ll$\textsf{#1 --IG}$\gg$}}
\newcommand{\kr}[1]{\textcolor{purple}{$\ll$\textsf{#1 --KR}$\gg$}}

\newcommand\footnoteref[1]{\protected@xdef\@thefnmark{\ref{#1}}\@footnotemark}

\newcommand{\ntv}{{node2vec}\xspace}
\newcommand{\Ntv}{{Node2vec}\xspace}
\newcommand{\dfs}{{tBDFS}\xspace}
\newcommand{\gcn}{{GCN}\xspace}
\newcommand{\debias}{{Debias}\xspace}
\newcommand{\fairgnn}{{FairGNN}\xspace}

\newcommand{\argmin}{\mathop{\mathrm{argmin}}} 
\def\eqd{\,{\buildrel d \over =}\,} 
\newcolumntype{L}[1]{>{\raggedright\let\newline\\\arraybackslash\hspace{0pt}}m{#1}}
\newcolumntype{C}[1]{>{\centering\let\newline\\\arraybackslash\hspace{0pt}}m{#1}}
\newcolumntype{R}[1]{>{\raggedleft\let\newline\\\arraybackslash\hspace{0pt}}m{#1}}
\newcommand\independent{\protect\mathpalette{\protect\independenT}{\perp}}
\def\independenT#1#2{\mathrel{\rlap{$#1#2$}\mkern2mu{#1#2}}}
\def\multitable#1{\multicolumn{1}{p{0.5cm}}{\centering #1}}

\def\PP{{\mathbb P}}
\def\EE{{\mathbb E}}
\def\RR{{\mathbb R}}
\def\II{{\mathbb I}}
\def\NN{\mathcal{N}}
\def\PP{\mathcal{P}}

\def\g{\mathbf{G}}
\def\F{F}
\def\D{D}
\def\E{\mathbf{E}}
\def\V{\mathbf{V}}
\def\X{\mathbf{X}}
\def\Y{\mathbf{Y}}
\def\A{\mathbf{A}}
\def\y{\mathbf{y}}
\def\a{\mathbf{a}}
\def\B{\mathbf{B}}
\def\h{\mathbf{h}}
\def\P{\mathbf{P}}
\def\Q{\mathbf{Q}}
\def\W{\mathbf{W}}

\newcommand{\change}[1]{\color{red}{#1 }\color{black}}
\begin{abstract}
Temporal graph neural networks (temporal GNNs) have been widely researched, reaching state-of-the-art results on multiple prediction tasks. A common approach employed by most previous works is to apply a layer that aggregates information from the historical neighbors of a node. Taking a different research direction, in this work, we propose \dfs\ -- a novel temporal GNN architecture. \dfs applies a layer that efficiently aggregates information from temporal paths to a given (target) node in the graph. For each given node, the aggregation is applied in two stages: (1) A single representation is learned for each temporal path ending in that node, and (2) all path representations are aggregated into a final node representation. Overall, our goal is not to add new information to a node, but rather observe the same exact information in a new perspective. This  allows our model to directly observe patterns that are path-oriented rather than neighborhood-oriented. This can be thought as a Depth-First Search (DFS) traversal over the temporal graph, compared to the popular Breath-First Search (BFS) traversal that is applied in previous works. We evaluate \dfs over multiple link prediction tasks and show its favorable performance compared to state-of-the-art baselines. To the best of our knowledge, we are the first to apply a temporal-DFS neural network.
\end{abstract}

\section{Introduction}
\label{sec:intro}

Graphs are ubiquitous, and nowadays, many data sources spanning over diverse domains such as the Web, cybersecurity, economics,  biology, and others are being modeled as a graph. Modeling a data source as a graph allows to detect rich structural patterns which are useful for a large variety of machine learning tasks, such as link-prediction and node classification~\cite{cai2018comprehensive}.
Learning on graphs is challenging, with an handful of different approaches that have been proposed over the recent years. Among others, \emph{Graph Neural Networks} (GNNs) have demonstrated state-of-the-art results over many different datasets and tasks \cite{kipf2016semi,hamilton2017inductive,velivckovic2017graph}.
GNNs allow to learn rich node (and edge) representations by applying neural network layers on the graph's structure such as convolution (CNNs) or recurrent (RNNs) neural networks~\cite{wu2020comprehensive}. 

While learning over graphs has been widely researched, many aspects still remain a challenge. Such aspects include, among others, scaling learning models to large graphs, extracting meaningful features from the graph's
complex structure; and the focus of our work: handling \emph{temporal graphs}. A temporal graph captures the evolution of networks over time, associating each edge between any two nodes in the graph with a timestamp. Using temporal GNNs, learning on the graph occurs over time, by applying the neural network layers according to the timeline in which the graph's topology has evolved~\cite{skardinga2021foundations}. 

Most existing GNNs~\cite{kipf2016semi,hamilton2017inductive,velivckovic2017graph}, and specifically temporal GNNs~\cite{xu2020inductive}, extract features from the graph structure by following a rule with the following question in mind: ``What have your neighbors been telling you?''. Practically, this rule is implemented by aggregating for each node in the graph the information from its neighbors. This aggregation is differential, making it a general layer that can be connected to any desired deep-learning architecture, such as stacking it one over the other.
While stacking the GNN layers enlarges the receptive field\footnote{The \emph{receptive field} of a given node in the graph is defined by the set of nodes in the graph that may influence its final representation. Therefore, the first layer in the stack observes for each node its first-order (direct) neighbors, the second one observes its second-order neighbors, etc.} of a node to nodes further in the graph, it still follows the above rule, making it hard to find patterns that may follow other rules.

In this work, we take a novel perspective, by exploring a different rule over temporal graphs, which aims to answer the following question: ``Tell me how did you receive this information?''.
Intuitively, while previous GNNs observe the graph in a \textit{Breadth-First} manner (i.e., observing neighbors at each layer), we present an algorithm that also observes the graph in a \textit{Depth-First} manner (i.e., observing paths at each layer).
While there are previous GNN works \cite{yang2021spagan,chen2021graph,lin2021metapaths,ying2021transformers} that recognize the importance of observing a path in the graph (and specifically in a DFS manner), they do not leverage the rich information of the path, nor can be trivially extended to temporal graphs. We hypothesize that, some tasks and datasets on temporal graphs hold patterns that are more DFS oriented rather than BFS oriented.
Traversing the graph in a DFS manner allows to capture the effective order in which information flows in the network from a source to its target. This is in comparison to a BFS traversal, which always assumes that information flows from the neighborhood first, which is not always true; For example, it might be possible in some cases that, a node along a path to the target may have multiple unrelated neighbors, and hence, may add an undesired noise to the target node's representation.   

\begin{figure}
    \centering
    \includegraphics[width=0.30\textwidth]{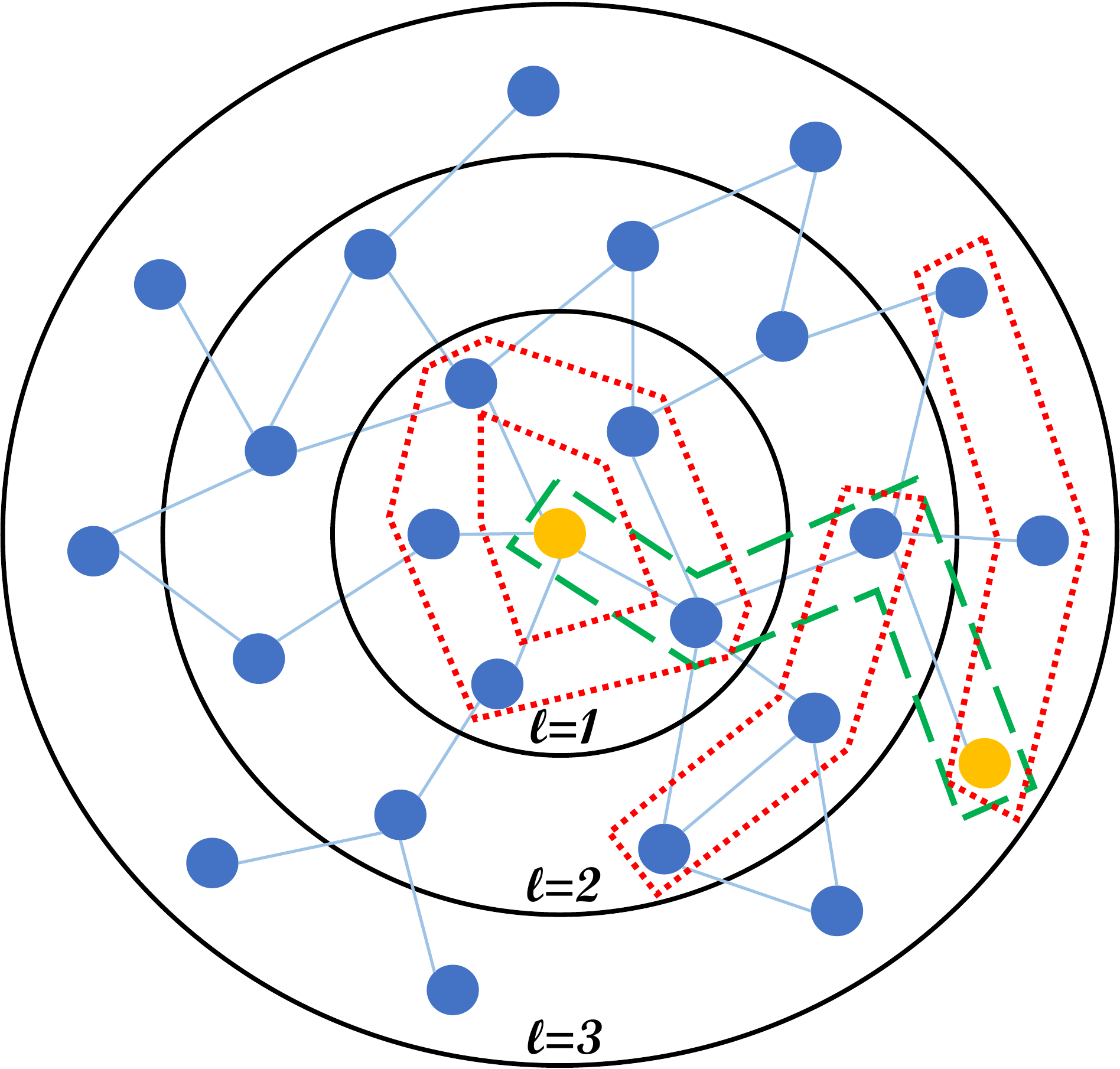}
    \caption{\label{fig:illustration} An illustration of how a given node (center) may depend on a distant node (bottom right). A BFS-oriented GNN is required to correctly propagate the information from the distant node through three other nodes, while also observing the ``unrelated'' neighbors (red-dotted area). Our DFS-oriented GNN captures the pattern directly via the path to that node (green dashed area).}
\end{figure}

Figure \ref{fig:illustration} further illustrates an example of a task with a DFS pattern. We notice that, while a BFS approach is challenged with the extraction of the signal (as the information should correctly propagate through 3 hops), a DFS approach is able to extract the signal directly from the path without requiring any information propagation.
As a more concrete example, let us consider the Booking dataset, which is one of the datasets being studied in our work. Lets assume that the event of a user's visit to a city is represented by a temporal edge in the graph. If that same city was already visited in the past by the same user (i.e., user-city-user), a BFS approach would have to propagate the previous user node to the next user node via the intermediate city node.
When propagating to an intermediate node, the propagation is shared across all its neighbors, meaning that it will be hard for the BFS approach to ``remember'' the relevant neighbor out of all the given neighbors. This issue gets even worse when the path is longer, as the information exponentially vanishes the larger the path is and the more neighbors they are.
By utilizing the DFS approach, the path of user-city-user is given explicitly to the model, without being required to propagate the user through the city.

Our DFS representation of a node is learned by two consecutive components:
(1)  Given a temporal path (as information travels in a ``chronological'' order in time) from a source node to a desired target node, the first component learns a path representation that is aware of all the nodes, edges, features and timestamps along that path.
(2) Given the representations of all the temporal paths to a given target node from the previous component, the second component is responsible for aggregating all the paths into a final node representation.

It is important to note that, the DFS representation is presented with the exact same information as the BFS representation, with the only difference being in the way it is presented to the network.
While previous works have enlarged the receptive field (which resulted with more information that is fed to the network), we demonstrate that, even without enlarging the receptive field, the same exact information can be fed to the network in a DFS way, enabling it to easily extract DFS patterns. Furthermore, we propose an efficient way to learn the DFS representation during the BFS learning phase.

Overall, we propose \dfs\ -- a temporal GNN architecture that given a node and a time, learns a DFS-aware representation of the node for that specific time.
In order to combine between a BFS representation (using previous works \cite{xu2020inductive}) and our proposed DFS representation, we suggest learning the balance between the two representations, resulting in a final node representation that is aware of both types. We empirically evaluate \dfs on a variety of link-prediction tasks using several real-world datasets, demonstrating its superior performance over competitive baselines. 

\section{Related Work}
\label{sec:rw}

We review related works along two main dimensions which are the most relevant to ours: learning over temporal graphs and works that have leveraged paths or DFS.

\subsection{Learning over Graphs}
The field of learning on graphs has been studied for many years, but has grown widely in the last few years. Approaches over the years can be viewed along three ``generarions'': matrix factorization, random walks, and graph neural networks.

The first generation included different factorization methods over the graph adjacency matrix. These works obtain node representations by preforming a dimensionality reduction over the adjacency matrix. This can be performed by learning to reconstruct the edges~\cite{Laplacian:NIPS2001,Yan:2007:GEE}, the neighbors~\cite{Roweis:nonlineardimensionality:science}, or even the entire k-hop neighbors \cite{cao2015grarep,tenenbaum:global:2000}.

The second generation proposed random walks over the graph in order to create a corpus representing the graph structure. For instance, DeepWalk~\cite{Perozzi:2014:DOL} represents the nodes as words and random walks as sentences, thereby reducing the problem to Word2Vec~\cite{word2vec}. Node2Vec~\cite{grover16node} extended this work by enabling a flexible neighborhood sampling strategy, which allows to smoothly interpolate between BFS and DFS.

The most recent generation of methods considers graph neural networks (GNN), which introduce generic layers that can be added to any desired deep learning architecture. For example, GCN~\cite{kipf2016semi} and GraphSAGE~\cite{hamilton2017inductive} presented a convolutional layer that computes the average neighbor representation. Graph attention networks (GAT)~\cite{velivckovic2017graph} present an attention-based technique that learns the importance of each neighbor to the central node. \cite{ying2018hierarchical} introduces a dedicated pooling mechanism for graphs that learns the soft clusters that should be pooled together. \cite{schlichtkrull2018modeling} introduced a method that can handle different types of relation edges. 
Many works have demonstrated the superiority of GNNs over different tasks, such as molecules property prediction \cite{gilmer2017neural}, protein-protein interaction prediction \cite{10.1093/bioinformatics/btaa1036}, fair job prediction \cite{aaai2022-eqgnn}, human movement \cite{jain2016structural,stgcn2018aaai,Feng:2018:DPH}, traffic forecasting \cite{yu2017spatio,cui2018high} and other urban dynamics \cite{Wang:2017:RRL}.
While GNNs are used as a layer that can be added to any architecture, some works proposed self-supervised techniques for learning node representations that can reconstruct the graph structure~\cite{kipf2016variational}.

\subsection{Learning over Temporal Graphs}
Learning over temporal graphs (or dynamic networks) was widely studied in recent years. Earlier works applied matrix factorization or other types of aggregations over the temporal dimension~\cite{dunlavy2011temporal,yu17temporally}. Others~\cite{nguyen2018continuous} learned continuous dynamic embeddings using random walks that follow ``chronological'' paths that could only move forward in time. Such a time-sensitive random-walk has been shown to outperform static baselines (e.g., node2vec~\cite{grover16node}).
More recent works utilized deep neural-networks. Among these works, \cite{singer2019node} learned static representations for each graph snapshot and proposed an alignment method over the different snapshots. The final node representations were then obtained by using an LSTM layer that learns the evolution of each node over time.
GNNs over temporal graphs was proposed in TGAT~\cite{xu2020inductive}, which extends GAT \cite{velivckovic2017graph} using the temporal dimension when aggregating the neighbors. For a given node, TGAT observes its neighbors as a sequence ordered by time of appearance. It then applies an attention layer that aggregates the information with temporal-awareness using a time encoder. Our \dfs method extends over TGAT by adding a new layer that is responsible to capture DFS patterns in a temporal graph.

\subsection{DFS in Graphs}
Common to most previously studied methods (both over static and temporal graphs) is that each layer learns to aggregate the neighbors of a target node (i.e., in a BFS manner). Compared to that, in this work, we take a different approach and propose a temporal GNN that aggregates information in a DFS manner along paths that end at the target node. 

Several previous works have further leveraged the ``DFS view'' of the graph.
Among these works, \cite{grover16node} have suggsted a random-walk approach with a flexible neighborhood sampling strategy between BFS and DFS exploration. However, their method does not handle node and edge features, nor can it be treated as a general differential layer. \cite{liu2019geniepath} have proposed to add a memory gate between the GNN layers, enabling to better remember how information has arrived to the target node. Yet, their method does not observe the DFS patterns, making it hard to extract specific paths that are important.
\cite{lin2021metapaths} have proposed a method for learning over heterogeneous graphs using \emph{metapaths}. Given a metapath, instances of it are sampled, while first-order nodes are aggregated separately from the higher-order ones.  Therefore,  their method as well does not actually leverage the DFS paths, but rather only perform aggregation of neighbors.
\cite{yang2021spagan} have further utilized a higher-order neighborhood by sampling nodes with shortest paths.
These nodes were expected to be more relevant to the target node.
Overall, none of the aforementioned works has actually leveraged the pattern of the path, nor handled temporal edges.
\cite{chen2021graph} also proposed to aggregate information from higher-order nodes. To this end, for each node, a representation was first learned using a GNN. Then, random paths were sampled, where the importance of the last node of each path to the target node was learned via an LSTM model. The final representation was obtained as the weighted sum of all the last nodes. However, this method did not leverage actual information from the path, but rather has only learned its weight.
Finally, \cite{ying2021transformers} have proposed to leverage the entire graph structure during the GNN aggregation, while weighting the importance of two nodes by their shortest path. Yet, as the authors testify, their approach could only handle very small graphs with few nodes, which does not scale well.

\subsection{Main Differences}
Our work differs from previous works in several ways.
 Firstly, we generalize to temporal graphs which enables us to learn how information dynamically travels in the graph.
Secondly, given a path, we learn both the importance of each of its nodes to its own representation; and the importance of each path to the target node representation.
Thirdly, we leverage node features, edge features, and timestamps.
Fourthly, we demonstrate the performance of the DFS aggregation on the exact same receptive field as the BFS aggregation. Differently from previous works, we show how actual DFS aggregation boosts performance, and not larger receptive fields. Furthermore, we sample the DFS paths recursively during the BFS aggregations, making it more efficient.

\section{\dfs Architecture}
\label{sec:framework}

In this section we present the main building blocks that allow to learn both BFS and DFS patterns in a temporal graph. We start with basic temporal graph notations. We then shortly describe BFS graph attention and then present our proposed alternative of DFS graph attention. Our \dfs approach is derived by combining both attention types.

Let $G=(V,E)$ now formally denote a temporal graph with nodes-set $V$ and edges-set $E$, respectively. We denote $(i,j)_t \in E$ as an undirected temporal edge, where node $i$ connects with a node $j$ at timestamp $t$.
We denote $x_i\in\RR^d$ the features of node $i\in V$, where $d$ is the number of features in the dataset.
We further denote $x_{i,j}(t)\in \RR^d$ the features of edge $(i,j)_t \in E$.
For a given prediction task defined by a given loss function, our goal now is to find for each node $i \in V$ and timestamp $t$ a feature-vector (representation) $h_i(t)\in \RR^d$ that minimizes the loss.

\subsection{Functional Time Encoding}
When using sequences in attention mechanisms, it is common to use positional encoding to allow the model to know the positions of the elements of the sequence \cite{kenton2019bert}. A problem raises when the sequence is continues rather than equally quantized (such as continues time series). In this case, positional encoding lacks to hold the continues information of the sequence. Therefore, an appropriate encoder is required. Instead of using positional embedding, we follow \cite{xu2020inductive} and leverage a continues time encoder $\Phi: T \rightarrow R^d$ from the time domain to a d-dimensional vector space:
\begin{equation}
\begin{split}
\label{eq:time_encoder}
\Phi(t) = \sqrt{\frac{1}{d}} \big[cos(w_1t), sin(w_1t),\dots,\\
cos(w_dt),sin(w_dt)\big]
\end{split}
\end{equation}
where $w_1,..w_d$ are trainable parameters of the model. This encoder holds special properties following Bochner’s Theorem \cite{loomis2013introduction}. We refer the reader to~\cite{xu2020inductive} for additional details.

\subsection{BFS Graph Attention}
\label{sec:bfs}

A common approach for learning temporal node representations $h_i(t)$, is to introduce a layer within a GNN setting that aggregates for each node the information from all its neighbors~\cite{kipf2016semi,hamilton2017inductive,velivckovic2017graph}. This can be thought of as a BFS aggregation, observing information in a \textit{Breadth-First} manner.

Following \cite{xu2020inductive}, given a timestamp $t$ and a node $i \in V$, with an initial feature vector $h^{(0)}_i=x_i$, a graph attention layer (at layer $l$) is used to update the node's features $h^{(l)}_i$ according to its neighborhood. To this end, we first mask out neighbors in the future (where $t_j$ denotes the timestamp in which an edge $(i,j)$ has been formed):
\begin{equation}
\begin{split}
\label{eq:temporal_neighbors}
\NN_i(t) = \{j \vert (i,j)_{t_j} \in E \wedge t_j<t\}
\end{split}
\end{equation}

We notice that, the same neighbor may appear twice in the past. Therefore, we refer to a specific appearance in time as a ``temporal neighbor''.
For each temporal neighbor $j \in \NN_i(t)$ that interacted with $i$ at timestamp $t_j<t$, we extract a feature representation that includes the neighbor's own embedding, the edge features, and a time difference embedding:
\begin{equation}
\begin{split}
\label{eq:neighbor_features}
h'^{(l-1)}_j = h^{(l-1)}_j(t_j) \Vert x_{i,j}(t_j) \Vert \Phi(\Delta t),
\end{split}
\end{equation}

where $\Delta t=t-t_j$, $\Phi$ is a time encoder as presented in Eq. \ref{eq:time_encoder}, which embeds a time difference into a feature vector of size $d$, and $\Vert$ is the concatenation operator. Similarly, for the target node $i$, we apply the same logic:
\begin{equation}
\begin{split}
\label{eq:self_features}
h'^{(l-1)}_i = h^{(l-1)}_i(t) \Vert \bar{0} \Vert \Phi(\Delta t),
\end{split}
\end{equation}

where $\bar{0} \in \RR^d$ is zero padding, as there is no actual edge between the target node to itself. Following~\cite{xu2020inductive}, we then apply a \emph{multihead cross-attention} as follows:
\begin{small}
\begin{equation}
\begin{split}
\label{eq:tgat}
\alpha^{m,(l)}_{ij}=\frac{exp(\W^{m,(l)}_Q h'^{(l-1)}_i \cdot \W^{m,(l)}_K h'^{(l-1)}_j)}{\sum_{j \in \NN_i(t)}{exp(\W^{m,(l)}_Q h^{(l-1)}_i \cdot \W^{m,(l)}_K h^{(l-1)}_j)}}, \\
h'^{(l)}_i(t) = {\Big\Vert}^M_{m=1} \left( \sum_{j \in \NN_i}{\alpha^{m,(l)}_{ij} \W^{m,(l)}_V h^{(l-1)}_j} \right), \\
h^{(l)}_i(t) = FFN(h^{(l-1)}_i(t) \Vert h'^{(l)}_i(t)),
\end{split}
\end{equation}
\end{small}

where $\W^{m,(l)}_Q$, $\W^{m,(l)}_K$, $\W^{m,(l)}_V \in \RR^{3d \times 3d}$ are trainable parameters of the model, $M$ is the number of attention heads, and $FFN$ is a feed-forward neural-network.

Stacking multiple GNN layers one over the other allows to enlarge the receptive field of the node. Let the number of stacked layers be $L$. The final node representation is then: 
$$h^{BFS}_i(t) = h^{(L)}_i(t)$$

\subsection{DFS Graph Attention}
\label{sec:dfs}

While the receptive field of a given node grows with more layers, the information is being propagated in a BFS way, where information is being aggregated by observing the node's neighbors (see Figure~\ref{fig:illustration}). Yet, by aggregating in such a manner, it is hard to extract information from a specific path. 
We therefore propose to learn an additional representation, $h^{DFS}_i(t)$, that observes the same receptive field as $h^{BFS}_i(t)$, but in a \textit{Depth-First} manner.

Given a target node (as illustrated in the center of Figure~\ref{fig:illustration}), stacking $L$ GNN layers provides us with a $L$-hop neighborhood that effects the BFS representation. A different way of observing this exact same neighborhood, would be by extracting all the paths of length $L$ ending at the target node.
Let $\PP^L_i(t)$ represent the group of all temporal paths of length $L$ ending at node $i$ at timestamp $t$.
A specific path, $p_r=(j_0, j_1, ..., j_L) \in \PP^L_i(t)$,  holds a list of all nodes in the path (ordered from the latest to the earliest in time), where $j_0=i$, and $j_L$ is the last (source) node in the path.
As we want to propagate the information in a \textit{Depth-First} way, we observe each path separately. In order to do so, we aggregate the nodes in the path $p_r$ into a single representation.
 We next note that, such a path may be thought as a sequence of edge formation events over time.  In this work, we have chosen to leverage the attention mechanism as it demonstrated state-of-the-art results over sequential data aggregation, as follows: 
\begin{equation}
\begin{split}
\label{eq:tdfs_path}
\alpha^{m,(l)}_{ij}=\frac{exp(\W^{m,(l)}_Q h'^{(l-1)}_i \cdot \W^{m,(l)}_K h'^{(l-1)}_j)}{\sum_{j \in p_r}{exp(\W^{m,(l)}_Q h^{(l-1)}_i \cdot \W^{m,(l)}_K h^{(l-1)}_j)}} \\
h'^{(l)}_{i,r}(t) = {\Big\Vert}^M_{m=1} \left( \sum_{j \in p_r}{\alpha^{m,(l)}_{ij} \W^{m,(l)}_V h^{(l-1)}_j} \right) \\
h^{(l)}_{i,r}(t) = FFN(h^{(l-1)}_{i,r}(t) \Vert h'^{(l)}_i(t))
\end{split}
\end{equation}

where $\W^{m,(l)}_Q$, $\W^{m,(l)}_K$, $\W^{m,(l)}_V \in \RR^{3d \times 3d}$ are trainable parameters of the model, $M$ is the number of attention heads, and $FFN$ is a feed-forward neural network.
We can notice that Eq.~\ref{eq:tdfs_path} is quite similar to Eq.~\ref{eq:tgat}. This is not by accident, as the latter performs a BFS aggregation of the neighbors, while the former performs a DFS aggregation over the path nodes. Overall, there are three main differences between the two: (1) The aggregation in Eq.~\ref{eq:tdfs_path} is over nodes in a specific path, while in Eq.~\ref{eq:tgat} it is over neighbors of a specific node. (2) While BFS uses the time difference of a node from its parent, DFS uses the time difference of each node in the path from the target node.
(3) The aggregation is per a single path of the target node. That actually means that, in order to obtain the target node's final representation, we still need to combine the representation of all its associated paths.

At this point, for each node $i\in{V}$, we hold a representation for each of its possible paths, where $i$ acts as the path target. Therefore, in order to obtain a single representation for node $i$, we further aggregate all its path representations, as follows:
\begin{equation}
\begin{split}
\label{eq:tdfs}
h^{(l)}_{i}(t) = Aggregate(\{h^{(l)}_{i,r}(t) \vert p_r \in \PP^L_i(t)\}),
\end{split}
\end{equation}

where the $Aggregate$ function can be any aggregation such as average or attention. In this work, we have chosen to leverage the \emph{multi-head attention} mechanism~\cite{vaswani2017attention}; where for each node $i$, we treat $h^{BFS}_i(t)$ as the query, and $\{h^{(l)}_{i,r}(t) \vert p_r \in \PP^L_i(t)\}$ as both the keys and values.

It is important to note that, for a receptive field of $L$-hops, the BFS method proposed in Section~\ref{sec:bfs} is required to learn a different attention layer for each hop (i.e., an attention layer for each GNN layer). Compared to that, our method requires only two layers for any given receptive field: one layer for aggregating the path into a single representation, and the second for aggregating all paths into a final node representation. This means that, we can define $h^{DFS}_i(t) = h^{(1)}_{i}(t)$ after only one layer, without having to stack layers in order to capture information from the desired receptive field.

\subsection{\dfs: Balancing BFS and DFS}
As the trade-off between BFS and DFS may vary among graphs in the real world, we apply a final aggregation over the two representations (deriving our overall \dfs approach):
\begin{equation}
\label{eq:bfs_dfs}
h'_i(t) = \alpha \cdot h^{BFS}_i(t) \\ + (1 - \alpha) \cdot h^{DFS}_i(t),
\end{equation}

where $\alpha \in [0, 1]$ is a hyper-parameter that is responsible of balancing and smoothly interpolating between BFS and DFS.

\subsection{Efficient Path Sampling}

Extracting $\PP^L_i(t)$ in a brute-force way can be very time consuming. As we need to also calculate the BFS representations (see Eq.~\ref{eq:bfs_dfs}), we further propose a way to capture the temporal paths during the BFS implementation, making it more efficient.
Since the BFS implementation is recursive (each layer is a deeper call that ``explodes'' the new neighbors), in every recursive call, we explode the current path up to node $i$ with its temporal neighbors, $\NN_i(t)$, into $|\NN_i(t)|$ new paths (that will continue to explode in the next recursive calls).
When the recursive call reaches its final depth ($L$), we notice that the group of exploded paths is equal to $\PP^L_i(t)$. This is done without any additional computation over the BFS method.
Next, all is left to do is to propagate the paths back to the initial call and run Eq.~\ref{eq:tdfs_path} and Eq.~\ref{eq:tdfs} on the paths in $\PP^L_i(t)$.
This, therefore, not just resolves us with the DFS node representation, but also promises that the DFS aggregation is presented with the exact same information as the BFS aggregation.

\section{Evaluation}
\label{sec:experiments}

\begin{table*}[t!]
\center
\scriptsize
\setlength{\tabcolsep}{0.17em}
\caption{\label{table:main} Main results. Boldfaced results indicate a statistically significant difference.}
\hspace*{-4.0em}
\begin{tabular}{l@{}lll@{\hskip 0.03in}lll@{\hskip 0.03in}lll@{\hskip 0.03in}lll@{\hskip 0.03in}lll@{}}
\toprule
\multicolumn{2}{c}{\multirow{2}{*}{Model}} & \multicolumn{2}{c}{Reddit} & & \multicolumn{2}{c}{Booking}  & & \multicolumn{2}{c}{Act-mooc}     & & \multicolumn{2}{c}{Movielens}  & & \multicolumn{2}{c}{Wikipedia}      \\ 
\cmidrule(lr){3-4}\cmidrule(lr){6-7}\cmidrule(lr){9-10}\cmidrule(lr){12-13}\cmidrule(lr){15-16}
\multicolumn{1}{l}{} & & Accuracy & F1      &  & Accuracy   & F1        &    & Accuracy  & F1            & & Accuracy  & F1   &  & Accuracy  & F1    \\ 
\midrule
\multirow{4}{*}{\rotatebox{90}{Temporal}} \hspace{1.0em}
&\dfs 	 & $\mathbf{68.70 (\pm 0.36)}$ & $\mathbf{74.04 (\pm 0.25)}$ &	 & $\mathbf{74.71 (\pm 0.51)}$ & $\mathbf{78.84 (\pm 0.52)}$ 	& & $\mathbf{56.45 (\pm 0.30)}$ & $69.15 (\pm 0.10)$ &	 & $\mathbf{72.94 (\pm 0.09)}$ & $72.76 (\pm 0.59)$ 	 & & $\mathbf{86.99 (\pm 0.15)}$ & $\mathbf{87.32 (\pm 0.16)}$\\
& TGAT 	 & $66.26 (\pm 0.13) $ & $72.23 (\pm 0.11) $  & 	 & $73.24 (\pm 0.29) $ & $78.06 (\pm 0.28) $  & 	 & $55.92 (\pm 0.03) $ & $69.08 (\pm 0.17) $  & 	 & $72.65 (\pm 0.06) $ & $72.84 (\pm 0.50) $  & 	 & $86.79 (\pm 0.05) $ & $87.03 (\pm 0.06) $\\
& tNodeEmbed 	 & $67.67 (\pm 0.89) $ & $68.59 (\pm 0.46) $ 	& & $58.85 (\pm 0.49) $ & $55.80 (\pm 0.80) $ 	 & & $54.66 (\pm 0.66) $ & $54.43 (\pm 1.30) $ 	& & $55.68 (\pm 1.06) $ & $38.88 (\pm 4.01) $ &	 & $67.32 (\pm 0.52) $ & $67.44 (\pm 0.31) $\\
& GAT+T 	 & $65.31 (\pm 0.55) $ & $73.22 (\pm 0.33) $  & 	 & $62.56 (\pm 0.27) $ & $65.00 (\pm 0.45) $  & 	 & $52.23 (\pm 0.58) $ & $66.48 (\pm 0.52) $  & 	 & $57.25 (\pm 0.90) $ & $61.32 (\pm 0.77) $  & 	 & $76.0 (\pm 0.59) $ & $79.52 (\pm 0.40) $  \\
\midrule[0.01em]
\multirow{3}{*}{\rotatebox{90}{Static}}
& VGAE 	 & $65.47 (\pm 0.23) $ & $72.35 (\pm 0.13) $  & 	 & $60.40 (\pm 0.32) $ & $62.22 (\pm 0.57) $  & 	 & $52.15 (\pm 0.20) $ & $66.55 (\pm 0.09) $  & 	 & $56.90 (\pm 0.31) $ & $62.68 (\pm 0.11) $  & 	 & $72.69 (\pm 0.17) $ & $77.64 (\pm 0.15) $  \\
& GAE 	 & $66.24 (\pm 0.17) $ & $72.75 (\pm 0.10) $  & 	 & $61.31 (\pm 0.14) $ & $62.63 (\pm 0.18) $  & 	 & $51.00 (\pm 0.48) $ & $66.57 (\pm 0.10) $  & 	 & $56.93 (\pm 0.24) $ & $62.67 (\pm 0.35) $  & 	 & $73.32 (\pm 0.11) $ & $77.94 (\pm 0.10) $  \\
& node2vec 	 & $59.85 (\pm 0.03) $ & $70.37 (\pm 0.03) $  & 	 & $60.94 (\pm 0.13) $ & $64.69 (\pm 0.18) $  & 	 & $49.47 (\pm 0.23) $ & $57.49 (\pm 0.18) $  & 	 & $51.94 (\pm 0.13) $ & $54.82 (\pm 0.23) $  & 	 & $69.85 (\pm 0.30) $ & $75.06 (\pm 0.27) $  \\
\bottomrule
\end{tabular}
\end{table*}

As a concrete task for evaluating \dfs, we now apply it on the link-prediction task over a variety of temporal graph datasets. We first describe the datasets and our experimental setup (model implementation and training, baselines and metrics). We then present the evaluation results. 

\subsection{Datasets}
The following datasets were used in our evaluation:

\begin{itemize}
\item \textbf{Wikipedia}~\cite{kumar2019predicting}: A bipartite graph representing users editing Wikipedia pages. Each edge represents an edit event with its timestamp. 

\item \textbf{Reddit}~\cite{kumar2018community}: A graph representing links between subreddits in the Reddit website. A link occurs when a post in one subreddit is created with a hyperlink to a post in a second subreddit.

\item \textbf{Act-mooc}~\cite{kumar2019predicting}: A bipartite graph representing students taking courses. Each edge is represented with a timestamp of when the student took the course.

\item \textbf{MovieLens}~\cite{harper2015movielens}: A bipartite graph representing $1$ million users' movie ratings. Each edge represents a rating event with its timestamp.

\item \textbf{Booking}~\cite{goldenberg2021booking}: A bipartite graph used in the WSDM2021 challenge. A temporal edge represents a user visiting a city at a given time.
\end{itemize}

 We split each dataset over time, with $70\%$ of the edges used for training, $15\%$ used for validation, and the rest (most recent) $15\%$ for testing.

\subsection{Experimental Setup}
\subsubsection{Model implementation and training}
We implement \dfs\footnote{GitHub repository with code and data: \url{https://github.com/urielsinger/tBDFS}} 
with pytorch~\cite{paszke2017automatic}. We use the Adam~\cite{kingma2015adam} optimizer, with a learning rate of $10^{-4}$, $\beta_1 = 0.9$, $\beta_2 = 0.999$, $\ell_2$, dropout $p=0.1$ and batch size of $200$.
For a fair comparison, following~\cite{kipf2016semi}, for all GNN baselines (including \dfs), we set the number of layers $L=2$.

During training, for each temporal edge $(i,j)_t$, we sample a negative edge $(i,j')_t$, and learn to contrast between the two, following the loss proposed in~\cite{xu2020inductive}:
\begin{equation}
\begin{split}
\label{eq:objective}
Loss = -\sum_i \big[ log(\sigma(FFN(h'_i(t) \Vert h'_j(t)))) \\ + log(\sigma(-FFN(h'_i(t) \Vert h'_{j'}(t)))) \big],
\end{split}
\end{equation}

where $\sigma(\cdot)$ is the sigmoid activation function, and $FFN$ is a feed-forward neural network.

During inference, we choose amongst the most likely link, where we calculate the link probability between two candidates as follows: $\sigma(FFN(h'_i(t) \Vert h'_j(t)))$.

\subsubsection{Baselines}
\begin{itemize}
\item \textbf{node2vec}~\cite{grover16node} is a common baseline for representation learning over graphs. Its core idea is to turn the graph into ``sentences'' by applying different random walks. These sentences are then used for training a word2vec~\cite{word2vec} model that resolves with a representation for each node.

\item \textbf{GAE}~\cite{kipf2016variational} is a Graph-Auto-Encoder model. The encoder consists of Graph-Convolutional-Network (GCN)~\cite{kipf2016semi} layers that resolves with a representation for each node. The decoder then tries to reconstruct the edges of the graph using the dot-product between node pairs.

\item \textbf{VGAE}~\cite{kipf2016variational} is a variational version of the GAE model.

\item \textbf{GAT+T}~\cite{velivckovic2017graph} is a state-of-the-art method for static GNNs. We adapt GAT by adding edge and time features. We further use the same time-encoder $\Phi$ presented in \cite{xu2020inductive} for the time features.

\item \textbf{tNodeEmbed}~\cite{singer2019node} learns a static representation for each static graph snapshot. An alignment is then applied over the snapshots to learn node representation between consecutive snapshots. An LSTM model is then trained to learn the final node representation by aggregating the various node (snapshot) representations over time.

\item \textbf{TGAT}~\cite{xu2020inductive} is currently the state-of-the-art method for temporal GNNs. This is a BFS only version of our method (i.e., $\alpha=1$, and implemented according to Section~\ref{sec:bfs}), and therefore carries special importance. 
\end{itemize}

\subsubsection{Evaluation metrics}
\label{sec:metrics}
We evaluate the performance of our model over the \emph{temporal link prediction} task. To this end, we treat existing temporal edges as ``positive edges''. We further randomly sample negative edges equally to the amount of the positive edges on each dataset.
We report the prediction Accuracy and F1-score (F1).
We report the average metrics over $5$ different seeds, and validate statistical significance of the results using a two-tailed paired Student's t-test for $95\%$ confidence.
\subsection{Main Results}
\label{sec:result_main}

\begin{table*}[tb]
\center
\scriptsize
\setlength{\tabcolsep}{0.17em}
\caption{\label{table:ablation} Ablation results. Starting from the second row, a single component is removed from the model.}
\begin{tabular}{@{}lll@{\hskip 0.03in}lll@{\hskip 0.03in}lll@{\hskip 0.03in}lll@{\hskip 0.03in}lll@{}}
\toprule
\multicolumn{1}{l}{\multirow{2}{*}{Model}} & \multicolumn{2}{c}{Reddit} & & \multicolumn{2}{c}{Booking}  & & \multicolumn{2}{c}{Act-mooc}     & & \multicolumn{2}{c}{Movielens}  & & \multicolumn{2}{c}{Wikipedia}      \\ 
\cmidrule(lr){2-3}\cmidrule(lr){5-6}\cmidrule(lr){8-9}\cmidrule(lr){11-12}\cmidrule(lr){14-15}
\multicolumn{1}{l}{} & Accuracy & F1      &  & Accuracy   & F1        &    & Accuracy  & F1            & & Accuracy  & F1   &  & Accuracy  & F1    \\ 
\midrule
\dfs 	 & $68.70 (\pm 0.36)$ & $74.04 (\pm 0.25)$ &	 & $74.71 (\pm 0.51)$ & $78.84 (\pm 0.52)$ 	& & $56.45 (\pm 0.30)$ & $69.15 (\pm 0.10)$ &	 & $72.94 (\pm 0.09)$ & $72.76 (\pm 0.59)$ 	 & & $86.99 (\pm 0.15)$ & $87.32 (\pm 0.16)$\\
-BFS 	 & $68.42 (\pm 0.46) $ & $73.92 (\pm 0.30) $  & 	 & $74.11 (\pm 0.37) $ & $78.53 (\pm 0.37) $  & 	 & $55.61 (\pm 0.16) $ & $69.00 (\pm 0.17) $  & 	 & $72.82 (\pm 0.10) $ & $72.65 (\pm 0.51) $  & 	 & $86.94 (\pm 0.12) $ & $87.09 (\pm 0.26) $\\
-DFS 	 & $66.26 (\pm 0.13) $ & $72.23 (\pm 0.11) $  & 	 & $73.24 (\pm 0.29) $ & $78.06 (\pm 0.28) $  & 	 & $55.92 (\pm 0.03) $ & $69.08 (\pm 0.17) $  & 	 & $72.65 (\pm 0.06) $ & $72.84 (\pm 0.50) $  & 	 & $86.79 (\pm 0.05) $ & $87.03 (\pm 0.06) $\\
path-avg 	 & $66.93 (\pm 0.34) $ & $72.54 (\pm 0.19) $  & 	 & $72.76 (\pm 0.45) $ & $77.73 (\pm 0.26) $  & 	 & $56.10 (\pm 0.27) $ & $68.91 (\pm 0.26) $  & 	 & $72.42 (\pm 0.19) $ & $72.31 (\pm 0.54) $  & 	 & $85.47 (\pm 0.60) $ & $85.92 (\pm 0.43) $\\
paths-avg 	 & $69.92 (\pm 0.64) $ & $74.24 (\pm 0.25) $  & 	 & $74.64 (\pm 0.67) $ & $78.79 (\pm 0.37) $  & 	 & $56.71 (\pm 0.54) $ & $68.05 (\pm 1.25) $  & 	 & $71.79 (\pm 0.24) $ & $70.40 (\pm 0.76) $  & 	 & $86.06 (\pm 0.20) $ & $86.32 (\pm 0.23) $\\
-time 	 & $61.67 (\pm 0.46) $ & $62.23 (\pm 0.59) $  & 	 & $54.32 (\pm 3.30) $ & $55.52 (\pm 3.97) $  & 	 & $48.18 (\pm 1.68) $ & $51.85 (\pm 3.19) $  & 	 & $60.32 (\pm 0.41) $ & $56.34 (\pm 1.44) $  & 	 & $72.24 (\pm 0.25) $ & $67.97 (\pm 0.63) $\\
\bottomrule
\end{tabular}
\end{table*}

We report the main results of our evaluation in Table~\ref{table:main}. We first notice that the temporal baselines (\dfs, TGAT, GAT+T, and tNodeEmbed) outperform the static baselines (node2vec, GAE, and VGAE). This indicates the importance of the temporal patterns in the datasets. As we can further observe, overall, \dfs outperforms all baselines over all datasets. Interestingly, on some datasets, \dfs has only a small margin of improvement over TGAT. We hypothesize that, this may be attributed to the fact that these datasets have less ``DFS patterns'', meaning that most of the signal can be observed via the ``BFS patterns''. Case in point is the relatively higher performance of \dfs on the Reddit dataset. Among all datasets, this dataset is the only one in our evaluation that is a general graph, while the rest are bipartite graphs. Therefore, this implies that \dfs has much more opportunities to leverage diverse DFS patterns that exist in this dataset.

\subsection{Balancing between BFS and DFS}
\begin{figure}[ht]
    \centering
    \subfloat[Reddit]
        {\includegraphics[width=0.22\textwidth]{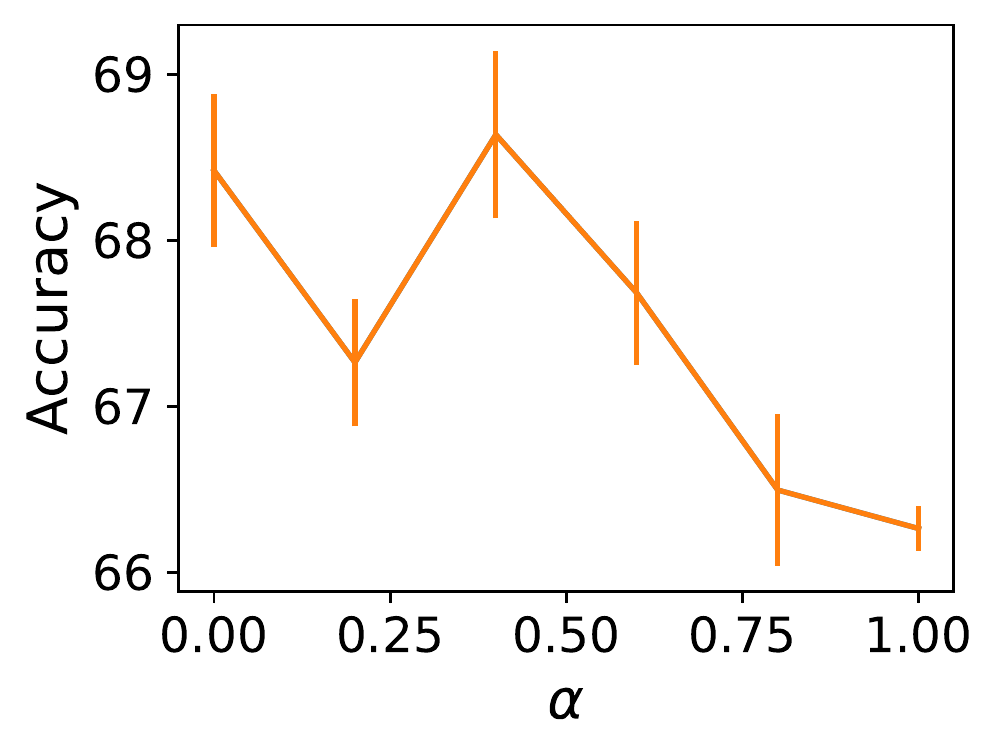}}
    \subfloat[Booking]
    	{\includegraphics[width=0.22\textwidth]{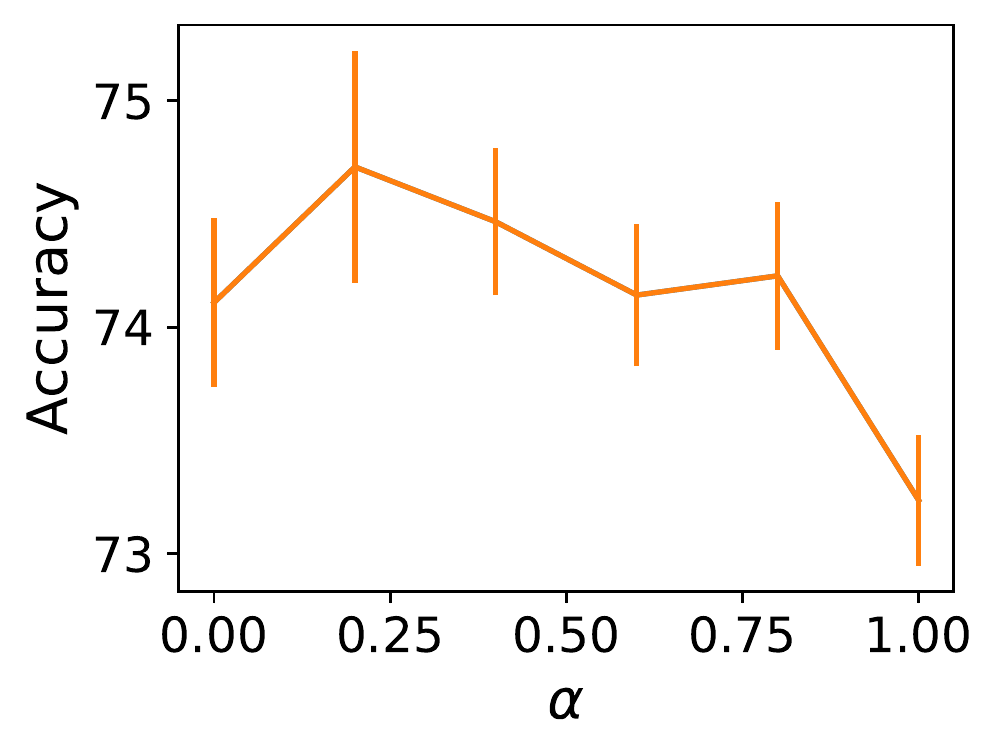}}
    \quad
    \subfloat[Act-mooc]
    	{\includegraphics[width=0.23\textwidth]{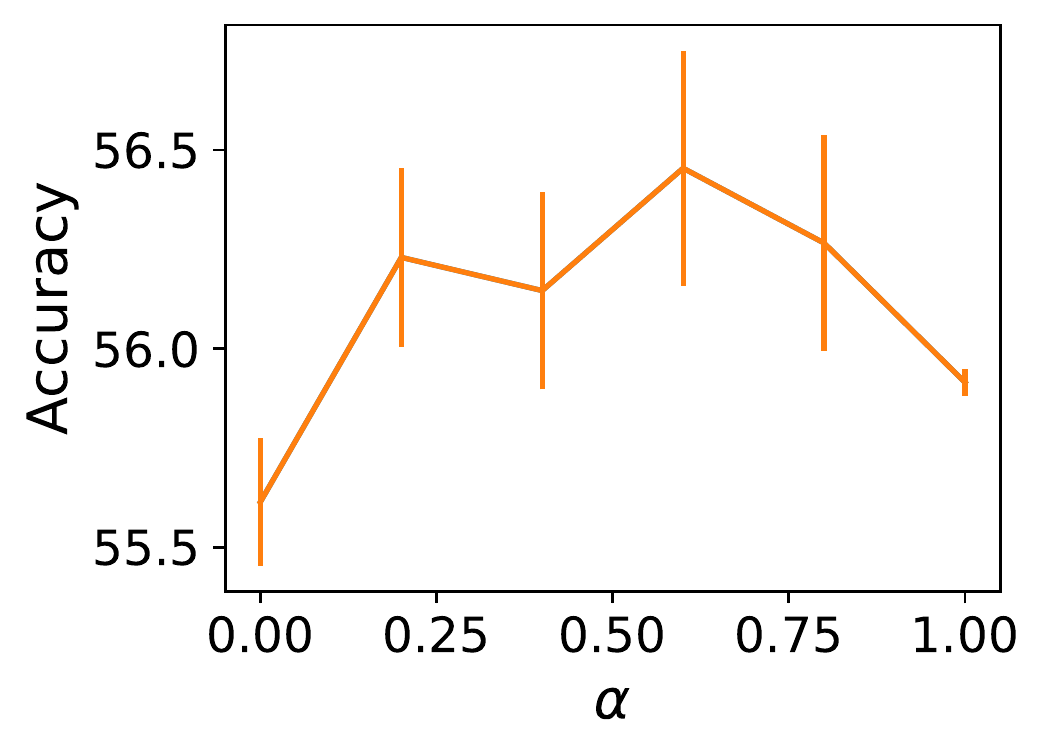}}
    \subfloat[Movielens]
    	{\includegraphics[width=0.23\textwidth]{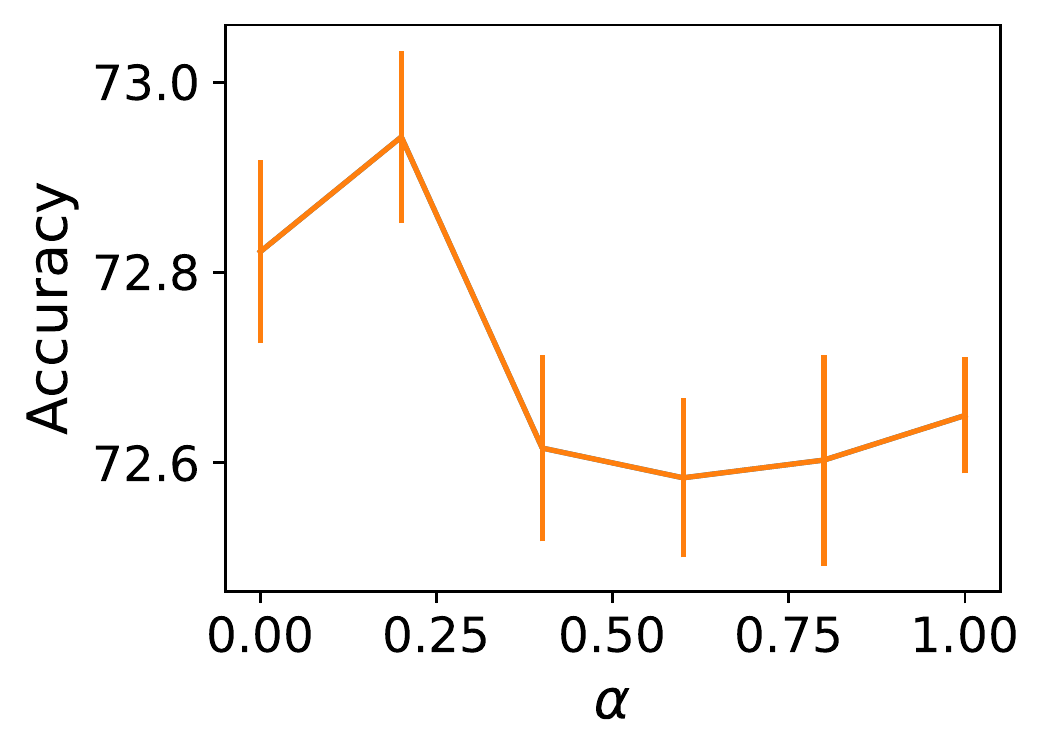}}
    \caption{\label{fig:bfs_dfs}Balancing between BFS and DFS. $\alpha=0$ means just DFS while $\alpha=1$ means just BFS.}
\end{figure}

We next analyze the importance of the $\alpha$ parameter over four datasets (the Wikipedia dataset has a similar trend; hence is omitted for space considerations).
As presented in Eq.~\ref{eq:bfs_dfs}, $\alpha$ is responsible for balancing and smoothly interpolating between BFS and DFS. We report in Figure~\ref{fig:bfs_dfs} the performance of \dfs using different values of $\alpha$ values in $[0,1]$. 
As we can observe, the best $\alpha$ is always a combination of BFS and DFS (i.e., $\alpha\in(0,1)$), and never one representation alone (i.e., $\alpha=0$ for DFS or $\alpha=1$ for BFS). This serves as a strong empirical evidence for the importance of augmenting the ``traditional'' BFS signal used in all previous works with the DFS-learned signal. 
We further observe that, except for Act-mooc, the DFS representation alone is better than the BFS representation alone. This strengthens our main hypothesis, which assumes that temporal graphs are more likely to follow a rule that aims to answer the ``DFS question'', i.e., ``Tell me how did you receive this information?''.

\subsection{Ablation Study}
We next report in Table~\ref{table:ablation} the results of our model's ablation study. To this end, we remove each time a single component from the \dfs model and measure its impact on performance. We explore a diverse set of ablations, as follows: 

\textbf{-DFS (-BFS)}: We remove the DFS (BFS) representation from the final node representation, ending up with merely a BFS (DFS) representation; setting $\alpha=1$ ($\alpha=0$).
As can be noticed, removing the BFS representation or the DFS representation, considerably degrades the performance.
This demonstrates that, there is no ``correct'' way to observe a graph, but rather by the combination of the two representations. This result was observed in many tasks on
graphs. Most relevant to our work, it was observed in
graph representation learning, e.g, ~\cite{grover16node} combined BFS and DFS walks during the random
walks.

\textbf{path-avg}: We switch the attention aggregation of the path, as presented in Eq. \ref{eq:tdfs_path}, to an average aggregation instead.
As we can observe, the performance degrades over all datasets and metrics. This demonstrates the importance of a smart aggregation layer over nodes in a given path.

\textbf{paths-avg}: We switch the paths attention aggregation, as presented in Eq. \ref{eq:tdfs}, to an average aggregation instead.
We observe that the performance degrades in 7 out of the 10 metrics, while additional 2 remain with similar performance.
We note that during the previous step (see Eq. \ref{eq:tdfs_path}), we could potentially learn that a given path is uninformative for the target node.
This implies that the second step only fine-tunes the path representations from the previous step;
explaining why this ablation is less effective than the ``path-avg'' ablation. 

\textbf{-time}: We remove the time information in two manners: (1) In the time encoder, for any given $\Delta t$, we set $\Phi(\Delta t)=\bar{0}$; (2) The future neighbors are not masked as explained in Eq. \ref{eq:temporal_neighbors}.
As we can see, removing the time information greatly degrades the performance also compared to the other temporal baselines. This reinforces the importance of dedicated temporal GNN architectures that leverage the temporal data.

\section{Conclusions}
\label{sec:conclusions}
We have explored a novel approach to observe temporal graphs. 
Most prior GNN works have proposed a layer that learns a node representation by aggregating information from its historical neighbors. 
Unlike prior GNN works, which have applied learning using a Breath-First Search (BFS) traversal over historical neighbors, we tackled the learning task from a different perspective and proposed a layer that aggregates over temporal paths ending at a given target node. The core idea of our approach lies in learning patterns using a Depth-First Search (DFS) traversal. Such a traversal has a better potential of explaining how messages have ``travelled'' in the graph until they have reached a desired target.  
The DFS representation was produced by first learning a representation for each temporal path ending in a given target node, and then aggregating all path representations into a final node representation. 
We empirically showed that \dfs method outperforms state-of-the-art baselines on the temporal link prediction task, over a variety of different temporal graph datasets.
To the best of our knowledge, we are the first to apply a temporal-DFS neural-network.
We do not add new information to a node, but rather observe the same information by a new perspective. As future work, we wish to explore the effect of longer temporal paths and additional perspectives aside from BFS and DFS.

\bibliography{tBDFS}

\end{document}